\documentclass[10pt,journal,compsoc]{IEEEtran}

\ifCLASSOPTIONcompsoc
  \usepackage[nocompress]{cite}
\else
  \usepackage{cite}
\fi

\usepackage{graphicx}
\usepackage{multirow}
\usepackage{amsmath,amssymb,amsfonts}
\usepackage{amsthm}
\usepackage{mathrsfs}
\usepackage{xcolor}
\usepackage{textcomp}
\usepackage{url}
\usepackage{booktabs}
\usepackage{algorithm}
\usepackage{algorithmicx}
\usepackage{algpseudocode}
\usepackage{listings}

\definecolor{projectpink}{RGB}{190,40,120}
\usepackage[
  colorlinks=true,
  urlcolor=projectpink
]{hyperref}

\begin{document}

\title{RobotPan: A 360$^\circ$ Surround-View Robotic Vision System for Embodied Perception}

\author{Jiahao~Ma$^{*}$,
        Qiang~Zhang$^{*}$,
        Peiran~Liu,
        Zeran~Su,
        Pihai~Sun,
        Gang~Han,
        Wen~Zhao,
        Wei~Cui,
        Zhang~Zhang,
        Zhiyuan~Xu,
        Renjing~Xu$^{\dagger}$,
        Jian~Tang$^{\dagger}$,
        Miaomiao~Liu$^{\dagger}$,
        and~Yijie~Guo%
\IEEEcompsocitemizethanks{%
\IEEEcompsocthanksitem $^{*}$Equal contribution.
\IEEEcompsocthanksitem $^{\dagger}$Corresponding author.
% \IEEEcompsocthanksitem Manuscript received xxx; revised xxx.
}}

% \markboth{Journal of \LaTeX\ Class Files,~Vol.~14, No.~8, August~2015}%
% {Author \MakeLowercase{\textit{et al.}}: RobotPan: A 360$^\circ$ Surround-View Robotic Vision System for Embodied Perception}
\markboth{IEEE TRANSACTIONS ON PATTERN ANALYSIS AND MACHINE INTELLIGENCE}
{Jiahao Ma \MakeLowercase{\textit{et al.}}: RobotPan: A 360$^\circ$ Surround-View Robotic Vision System for Embodied Perception}

\IEEEtitleabstractindextext{%

% \begin{center}
% \small\textit{Project page:} \url{https://robotpan.github.io/}
% \end{center}
% \vspace{0.25em}
\begin{center}
\small\textit{Project website:} \url{https://robotpan.github.io/}\par
\end{center}
\vspace{0.3em}

\begin{abstract}
Surround-view perception is increasingly important for robotic navigation and loco-manipulation, especially in human-in-the-loop settings such as teleoperation, data collection, and emergency takeover. However, current robotic visual interfaces are often limited to narrow forward-facing views, or, when multiple on-board cameras are available, require cumbersome manual switching that interrupts the operator's workflow. Both configurations suffer from motion-induced jitter that causes simulator sickness in head-mounted displays. We introduce a surround-view vision system deployed on the \emph{Tiangong 3.0} humanoid platform, combining six cameras with LiDAR for calibrated, synchronized, full 360° perception. We further present \textsc{RobotPan}, a feed-forward framework that predicts \emph{metric-scaled} and \emph{compact} 3D Gaussians from calibrated sparse-view inputs for real-time rendering, reconstruction, and streaming. \textsc{RobotPan} lifts multi-view features into a unified spherical coordinate representation and decodes Gaussians using hierarchical spherical voxel priors, allocating fine resolution near the robot and coarser resolution at larger radii to reduce computational redundancy without sacrificing fidelity. To support long sequences, our online fusion updates dynamic content while preventing unbounded growth in static regions by selectively updating appearance. Finally, we release a multi-sensor dataset tailored to 360$^\circ$ novel view synthesis and metric 3D reconstruction for robotics, covering navigation, manipulation, and locomotion on real platforms. Experiments show that \textsc{RobotPan} achieves competitive quality against prior feed-forward reconstruction and view-synthesis methods while producing substantially fewer Gaussians, enabling practical real-time embodied deployment.
\end{abstract}

}

\maketitle

\IEEEdisplaynontitleabstractindextext
\IEEEpeerreviewmaketitle

\begin{figure*}[t]
  \centering
  \includegraphics[width=0.85\textwidth]{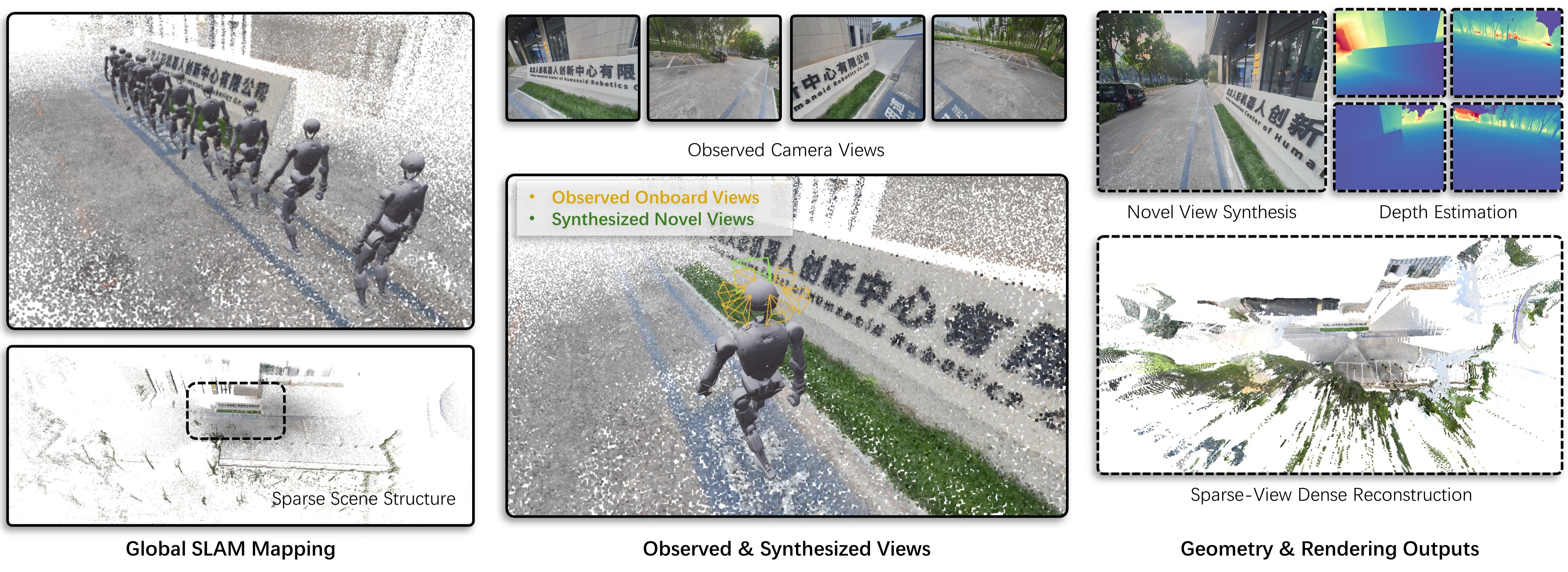}
  \caption{\textbf{System of RobotPan:} surround-view robotic vision system for real-time embodied perception which is deployed on the \emph{Tiangong 3.0} humanoid platform. Our system combines six cameras and LiDAR to provide full $360^\circ$ visual coverage for embodied robot operation. From calibrated sparse multi-view observations, \textsc{RobotPan} predicts metric-scaled and compact 3D Gaussians, enabling real-time surround-view rendering, novel view synthesis, depth estimation, and sparse-view dense reconstruction. 
By jointly supporting geometric consistency, compact representation, and real-time performance, the system serves as a practical robot visual interface for teleoperation, navigation, and loco-manipulation.}
  \label{fig:system_overview}
\end{figure*}

Robotic perception is a critical input for navigation, mobile manipulation, and embodied interaction, where vision provides rich geometric and semantic cues for autonomous decision-making and human-in-the-loop operation.
In practice, many robotic systems rely on monocular or stereo cameras, or camera--LiDAR fusion, due to their favorable trade-offs in cost, weight, power, and deployment complexity.
Such configurations are relatively easy to integrate on diverse platforms, scale well to different environments.

Despite these advantages, current robotic visual interfaces often fall short in human-in-the-loop operation, particularly during teleoperation, data collection, and emergency takeover.
A common workflow is that an operator wears a head-mounted display (HMD) to receive the robot's egocentric visual stream and issue control commands.
However, human vision and robot sensor layouts differ substantially, leading to several practical limitations.
First, the limited field of view of a forward-facing camera restricts situational awareness: operators cannot reliably perceive surrounding obstacles or affordances, which constrains safe motion and often requires uncomfortable body/head rotations to ``search'' for viewpoints.
Second, unlike the human vestibulo-ocular system that stabilizes perception during locomotion, robot-induced jitter and motion blur can make the HMD view unstable, causing simulator sickness and limiting the duration of effective operation.
Third, modern robots frequently mount multiple cameras for different purposes (e.g., head-mounted navigation cameras and body-mounted cameras for foot placement), but these views are typically isolated; switching among them adds cognitive load and interrupts the data-collection and control loop.

To address these issues, we propose a surround-view robotic vision system that combines six cameras with LiDAR to provide full 360$^\circ$ visual coverage.
This design, however, introduces challenges beyond those of panoramic cameras.
Unlike multi-fisheye panoramic rigs that approximate a single optical center, our cameras cannot be co-located due to hardware constraints, which makes classical image-stitching pipelines inapplicable and requires principled geometry-aware view synthesis.
Meanwhile, for downstream tasks such as SLAM and closed-loop control, the reconstructed scene must be metrically consistent.
Finally, the system must operate under real-time constraints with limited bandwidth and storage, demanding a compact 3D representation that supports both novel view synthesis and 3D reconstruction.

We present \textsc{RobotPan}, a feed-forward framework that predicts \emph{metric-scaled} and \emph{compact} 3D Gaussians from calibrated \textit{sparse} views input, enabling real-time rendering, reconstruction, and streaming. An overview of the system is illustrated in Fig.~\ref{fig:system_overview}.
Our key idea is to lift multi-view features into a unified \emph{spherical coordinate} and to decode Gaussians through \emph{hierarchical spherical voxel priors}.
The hierarchy allocates fine spatial resolution near the robot---where details matter for interaction---and progressively coarser resolution at larger radii, where a smaller number of larger Gaussians suffices to represent distant structure.
This distance-aware allocation yields a compact Gaussian set while preserving rendering fidelity, reducing redundancy substantially compared to dense pixel-wise Gaussian parameterizations.
Moreover, our per-frame predictions can be fused online to support long sequences: dynamic content is updated continuously, while static regions are preserved without unbounded Gaussian growth by selectively updating appearance (rather than spawning new primitives), significantly reducing memory overhead during streaming.

To facilitate research in embodied surround-view perception, we further introduce the first multi-sensor dataset tailored to 360$^\circ$ novel view synthesis and metric 3D reconstruction for robotics, covering navigation, manipulation and locomotion scenarios on real platforms.
Extensive experiments on both our proposed multi-sensor dataset and established public benchmarks demonstrate that \textsc{RobotPan} achieves competitive or even state-of-the-art quality against prior feed-forward reconstruction and view-synthesis methods, while producing markedly fewer Gaussians and lower computational redundancy, making it practical for real-time embodied deployment.
The contribution of this work are summarized as follows.
\begin{itemize}
  \item \textsc{RobotPan}, a feed-forward pipeline that predicts metric-scaled, compact 3D Gaussians from calibrated sparse views via hierarchical spherical priors, enabling real-time, streamable 360$^\circ$ surround-view rendering and metric-scaled reconstruction;
  \item A streaming Gaussian update strategy that incrementally fuses \emph{multi-frame, multi-camera} predictions, reduces redundancy, updates dynamic content, and prevents unbounded growth in static regions;
  % \item A spherical multi-camera--LiDAR robotic vision system that supports real-time 360$^\circ$ perception for human-in-the-loop and autonomous operation;
  \item A spherical multi-camera--LiDAR robotic vision system deployed on the \emph{Tiangong 3.0} humanoid platform, delivering real-time 360$^\circ$ perception for human-in-the-loop and autonomous operation;
  \item A new multi-sensor dataset and benchmark for embodied surround-view novel view synthesis and metric 3D reconstruction, designed to support a wide range of downstream robotic applications.
\end{itemize}
\section{Related Works}\label{sec:related_works}
\subsection{Feed-Forward 3D Reconstruction}
Recent learning-based methods enable end-to-end joint estimation of 3D geometry and camera parameters~\cite{dust3r,mast3r,reloc3r,spann3r}, avoiding the error accumulation of traditional multi-stage pipelines (feature matching, mapping, and triangulation). DUSt3R~\cite{dust3r} pioneered large-scale pretraining for feed-forward 3D reconstruction by regressing dense point maps from image pairs, avoiding explicit camera calibration. 
MASt3R~\cite{mast3r} strengthens correspondence quality on top of this paradigm, while Reloc3r~\cite{reloc3r} directly regresses relative 6-DoF poses and performs lightweight motion averaging for fast and accurate visual localization. A key limitation of pairwise processing is that inference can scale quadratically with sequence length.
To lift this constraint, multi-image variants introduce memory-based designs that aggregate context without exhaustive pairing, e.g., external spatial memory in Spann3R~\cite{spann3r} and multi-view modeling in MUSt3R~\cite{must3r}. 
For longer videos, SLAM3R~\cite{slam3r} adopts a sliding-window strategy for real-time dense reconstruction, and CUT3R~\cite{cut3r} maintains a persistent state to continuously update point maps in a common coordinate system. Beyond specialized pipelines, VGGT~\cite{vggt} unifies multi-view geometry prediction by directly inferring key 3D attributes (e.g., camera parameters, depth/point maps) from one to hundreds of views, and Fast3R~\cite{fast3r} further targets faster large-scale reconstruction. 
Several works couple such geometric priors with 3DGS~\cite{gs} for single-stage reconstruction and rendering, including MV-DUSt3R+~\cite{mvdust3r} and FLARE~\cite{flare}. 
Recent follow-ups push scalability and robustness further, e.g., permutation-equivariant, reference-free design in $\pi^3$~\cite{pi3}, test-time adaptation for better length generalization in TTT3R~\cite{ttt3r}, and training-free token merging acceleration for VGGT in FastVGGT~\cite{fastvggt}.

\subsection{Generalizable Novel View Synthesis}
NeRF~\cite{nerf} and Gaussian Splatting~\cite{gs} have significantly
advanced novel view synthesis~\cite{pixelnerf,mvsnerf,hashpoint}
and 3D reconstruction~\cite{OctreeGS,Scaffold-GS,MipSplatting,puzzles,dchm}, but their per-scene optimization limits real-time use. In contrast, \emph{generalizable} NVS learns a feed-forward prior from large multi-scene datasets to infer scene representations from a few context views at test time, and can be broadly grouped into generalized radiance fields~\cite{geonerf,pixelnerf,mvsnerf} and
feed-forward Gaussian splatting~\cite{pixelsplat,mvsplat,freesplat,
streamgs,salon3r,noposplat,splatt3r,anysplat,depthsplat,flare}. Among the former, PixelNeRF~\cite{pixelnerf} lifts pixel-aligned features, MVSNeRF~\cite{mvsnerf} incorporates multi-view stereo cues, and GeoNeRF~\cite{geonerf} injects explicit geometry priors; however, they typically assume known poses and struggle on long unposed streams. Recent generalizable 3DGS methods predict Gaussians feed-forward from sparse views: PixelSplat~\cite{pixelsplat} from image pairs, MVSplat~\cite{mvsplat} via cost-volume matching, and GS-LRM~\cite{gslrm} through a transformer-style large reconstruction model. For long sequences, SaLon3R~\cite{salon3r} employs compact anchor primitives with saliency-aware quantization to reduce Gaussian redundancy.

\subsection{3DGS with Quantization}
To enhance the practical applicability of 3D Gaussian Splatting, storage efficiency becomes a critical consideration. Recent approaches that compress and accelerate 3D Gaussians Splatting focus on codebook-based quantization, anchor or voxel abstractions, and hierarchical space partitioning to balance compactness and fidelity. CompGS~\cite{compgs} introduces a hybrid primitive design and compressed attribute representation, reporting large size reduction while maintaining quality. LightGaussian~\cite{lightgaussian} combines vector quantization with knowledge distillation and pseudo-view augmentation to improve compactness under distribution shift. Scaffold-GS~\cite{Scaffold-GS} adopts anchorized voxel structures to derive Gaussian attributes from sparse anchors, enabling view-adaptive rendering with fewer primitives. Octree-GS~\cite{OctreeGS} organizes Gaussians with an LOD-aware octree, improving consistency and real-time rendering via hierarchical culling and level selection. Together, these methods suggest that spatially structured anchors plus learned, lightweight decoders yield a promising path toward compact, scalable 3DGS for long videos and large scenes.

% \subsection{Panoramic Perception for Legged and Humanoid Robots.}
% 3D semantic occupancy, originating in autonomous driving with
% voxel-based~\cite{cao2022monoscene,wei2023surroundocc} and
% object-centric~\cite{huang2023tpv,huang2024gaussianformer}
% representations, has recently been extended to embodied
% agents~\cite{wu2024embodiedocc,zhang2025roboocc,kim2025mobileocc}.
% However, these methods inherit forward-facing, wheeled-platform
% assumptions, while legged and humanoid robots demand $360^\circ$
% coverage under self-occlusion, gait jitter, and tight payload
% budgets. Recent efforts address individual facets of this problem:
% spherical camera-LiDAR fusion~\cite{zhang2025humanoidpano},
% quadruped multimodal occupancy~\cite{zhao2026panommocc},
% vision-only panoramic SSC~\cite{shi2025oneocc}, end-to-end LiDAR
% collision avoidance~\cite{wang2025omniperception}, panoramic data
% synthesis~\cite{wu2025quadreamer,ni2025pano360}, and full-stack
% humanoid occupancy systems~\cite{cui2025humanoidoccupancy}.
% RobotPan builds on this line of work and takes a step further
% toward a generalized, deployable $360^\circ$ surround-view vision
% system that is practical across robot platforms.

\subsection{Panoramic Perception for Legged and Humanoid Robots.}
3D semantic occupancy, originating in autonomous driving with
voxel-based~\cite{cao2022monoscene,wei2023surroundocc} and
object-centric~\cite{huang2023tpv,huang2024gaussianformer}
representations, has recently been extended to embodied
agents~\cite{wu2024embodiedocc,zhang2025roboocc,kim2025mobileocc}
for richer spatial reasoning. However, these methods
inherit forward-facing, wheeled-platform assumptions that break
down on legged and humanoid robots, which demand $360^\circ$
coverage under self-occlusion, gait-induced ego-motion, and tight
onboard compute budgets. Recent efforts address individual facets
of this problem: spherical camera-LiDAR fusion~\cite{zhang2025humanoidpano},
quadruped multimodal occupancy with jitter
compensation~\cite{zhao2026panommocc}, vision-only panoramic
SSC via dual-projection fusion~\cite{shi2025oneocc}, end-to-end
LiDAR collision avoidance~\cite{wang2025omniperception}, panoramic
data synthesis~\cite{wu2025quadreamer,ni2025pano360}, and
full-stack humanoid occupancy systems~\cite{cui2025humanoidoccupancy}.
Building on this line of work, RobotPan takes a step further
toward a generalized, deployable $360^\circ$ surround-view vision
system that is practical across diverse robot platforms.
\section{Methodology}

\begin{figure*}[t]
  \centering
  \includegraphics[width=0.92\linewidth]{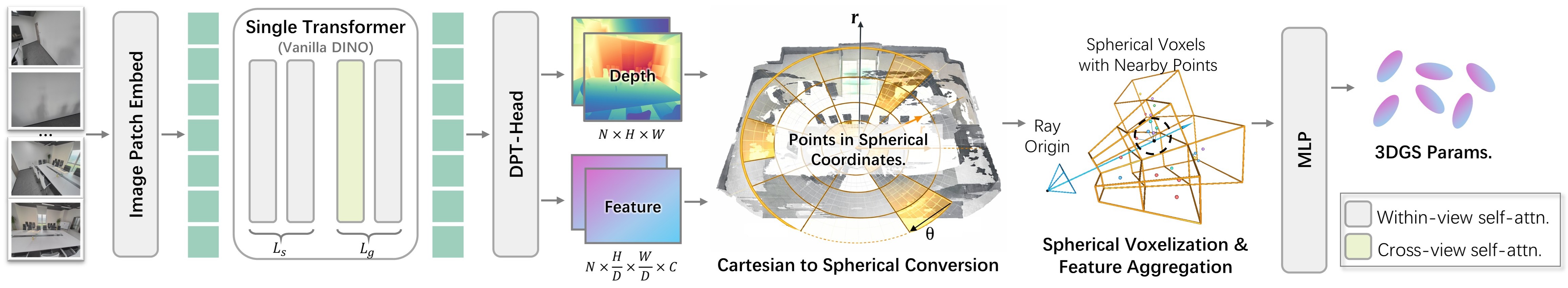}
  \caption{\textbf{Pipeline of RobotPan.} Multi-view images are encoded by a transformer to predict per-view depth and features. The reconstructed 3D points are converted to robot-centric spherical coordinates, voxelized into hierarchical spherical cells, and aggregated into anchor features, which are finally decoded into compact 3D Gaussian parameters for real-time rendering and reconstruction. }
  \label{fig:pipeline}
\end{figure*}

\subsection{Overview}
\label{sec:method_overview}
Our method has two components: (i) \textsc{RobotPan}, a single-pass feed-forward framework for 3D reconstruction (Sec.~\ref{sec:geometry_estimation}) and novel view synthesis (Sec.~\ref{sec:aggregation} to~\ref{sec:model_train}) from a \emph{time-synchronized} multi-camera rig, and (ii) a streaming fusion strategy that incrementally updates a unified, multi-view consistent 3D Gaussian Splatting (3DGS) representation over time (Sec.~\ref{sec:streaming}). 
At each time step $t$, the robot provides $V$ calibrated views $\{(I_i^t, P_i)\}_{i=1}^{V}$, where $I_i^t \in \mathbb{R}^{H \times W \times 3}$ and $P_i = K_i [R_i \mid \mathbf{t}_i]$ is the camera projection matrix defined by intrinsics $K_i$, rotation $R_i$, and translation $\mathbf{t}_i$.
We learn a feed-forward mapping
\begin{equation}
g_{\phi}:\{(I_i^t,P_i)\}_{i=1}^{V}\ \rightarrow\ \{(\boldsymbol{\mu}_j^t,\alpha_j^t,\Sigma_j^t,\mathbf{c}_j^t)\}_{j=1}^{N},
\label{eq:overview_mapping}
\end{equation}
where each Gaussian is parameterized by mean $\boldsymbol{\mu}$, opacity $\alpha$, covariance $\Sigma$, and color $\mathbf{c}$ (spherical harmonics). 
Unlike pixel-wise Gaussian prediction~\cite{pixelsplat, mvsplat}, \textsc{RobotPan} decodes Gaussians from \emph{hierarchical spherical anchor features}, producing a metric-scaled and compact set with reduced redundancy while maintaining rendering quality. 
Over a sequence, the synchronized observations form a multi-camera video stream $\{\{(I_i^t,P_i)\}_{i=1}^{V}\}_{t=1}^{T}$. 
We therefore employ a streaming Gaussian update strategy that fuses the \emph{multi-frame, multi-camera} predictions into a single global GS: it continuously refreshes dynamic content and prevents unbounded growth in static regions by selectively updating existing Gaussians (e.g., appearance) instead of na\"{\i}vely accumulating new ones.

\subsection{Metric-Scaled Multi-View Geometry Estimation}
\label{sec:geometry_estimation}
At each time step $t$, our robotic vision system receives $V$ calibrated images.
The cameras are mounted on a circular rig facing outward; their optical centers are not co-located.
In our setup, $V{=}6$ cameras are arranged with three front-facing and three rear-facing views, and the overlap between adjacent cameras is limited.

\textsc{RobotPan} maps the multi-view inputs to per-view geometric predictions in the \emph{camera coordinate} of each view:
\begin{equation}
f_{\theta}:\{(I_i^t,P_i)\}_{i=1}^{V}\ \rightarrow\ \{(\mathbf{X}_i^t,\ \mathbf{S}_i^t,\ \mathbf{F}_i^t)\}_{i=1}^{V},
\label{eq:geom_mapping}
\end{equation}
where $\mathbf{X}_i^t \in \mathbb{R}^{H \times W \times 3}$ is a metric-scaled point map,
$\mathbf{S}_i^t \in [0,1]^{H \times W}$ is a confidence map,
and $\mathbf{F}_i^t \in \mathbb{R}^{H \times W \times C}$ is a dense feature map.
Here, $(H,W)$ denotes the output resolution (typically the token/patch grid).
These geometric outputs serve as intermediate representations and are later consumed by our Gaussian prediction module.

As shown in Fig.~\ref{fig:pipeline}, we follow the overall architecture of VGGT~\cite{vggt}: a DINOv2~\cite{dinov2} backbone extracts per-image patch tokens, which are then processed by a transformer with alternating view-wise self-attention
(within each view) and global self-attention (across all views) to exchange information under limited overlap.
A lightweight decoder produces $(\mathbf{X}_i^t,\mathbf{S}_i^t,\mathbf{F}_i^t)$.
Motivated by $\pi^3$~\cite{pi3}, we treat the $V$ views as an unordered set and discard order-dependent components, including view-/frame-specific
positional embeddings and specialized learnable tokens that designate a reference view.
Camera calibration enters the model only through per-view geometric conditioning (via $P_i$), preserving permutation equivariance.
Finally, to ensure metric consistency, we fine-tune the geometry predictor with LiDAR supervision, aligning the predicted point maps to
LiDAR-derived metric geometry in each camera frame; training details are provided in Sec.~\ref{sec:model_train}.

\subsection{Spherical Voxel Feature Aggregation}
\label{sec:aggregation}

Pixel-wise Gaussian prediction is costly for real-time streaming, since it produces a dense set of primitives whose storage and bandwidth scale with image resolution.
To reduce redundancy, we introduce a \emph{camera-centered spherical voxel} representation that aggregates multi-view observations into a compact set of \emph{spherical anchor features}.
Spherical voxels naturally form a hierarchy: as the radius increases, the voxel volume grows, so distant regions can be represented by fewer, larger primitives; near-field regions, where interaction requires fine details, retain higher spatial resolution and yield smaller primitives (Fig.~\ref{fig:pipeline}, voxelization and feature aggregation stage).
In our experiments, this hierarchical design reduces storage by about $70\%$ compared to pixel-wise Gaussian prediction while preserving rendering quality.

\textbf{Spherical voxelization.}
The decoder outputs a per-view point map $\mathbf{X}_i^t \in \mathbb{R}^{H \times W \times 3}$.
For any pixel $p$, we denote the corresponding 3D point by $\mathbf{x}=\mathbf{X}_i^t(p)\in\mathbb{R}^3$.
After transforming $\mathbf{x}$ into the robot-centric frame using the known camera extrinsics $(R_i, \mathbf{t}_i)$, we then convert $\mathbf{x}=(x,y,z)$ to spherical coordinates $(r,\theta,\phi)$:
\begin{equation}
r=\|\mathbf{x}\|_2,\quad
\theta=\mathrm{atan2}(y,x),\quad
\phi=\mathrm{atan2}\!\left(z,\sqrt{x^2+y^2+\epsilon}\right).
\label{eq:spherical_coord}
\end{equation}
We discretize them into a spherical grid with bounds $r\in[r_{\min},r_{\max})$ and bin sizes $(\Delta r,\Delta\theta,\Delta\phi)$.
Each valid point is assigned to a voxel index
\begin{equation}
i_r=\left\lfloor\frac{r-r_{\min}}{\Delta r}\right\rfloor,\quad
i_\theta=\left\lfloor\frac{\theta-\theta_0}{2\pi}\,N_\theta\right\rfloor,\quad
i_\phi=\left\lfloor\frac{\phi-\phi_0}{\Delta\phi}\right\rfloor,
\label{eq:spherical_voxel}
\end{equation}
where $N_\theta=\lfloor 2\pi/\Delta\theta \rfloor$, and $(\theta_0,\phi_0)$ denote the lower bounds of the angular ranges.
This yields sparse voxel coordinates $\mathbf{C}^t=\{(i_r,i_\theta,i_\phi)\}$ for all valid points.

\textbf{Voxel volume analysis.}
Unlike uniform Cartesian voxels with constant volume, spherical voxels expand with radius.
For a voxel spanning $[r,r+\Delta r]$, $[\theta,\theta+\Delta\theta]$, and $[\phi,\phi+\Delta\phi]$, its volume can be approximated as
\begin{equation}
V(r,\theta,\phi)\ \approx\ r^2 \cos\phi\ \Delta r\,\Delta\theta\,\Delta\phi.
\label{eq:sph_voxel_volume}
\end{equation}
Thus, voxel volume grows roughly quadratically with $r$.
As a result, near-field space is partitioned into many small voxels, while far-field space is grouped into fewer large voxels that contain more points.
Since we regress a fixed number of Gaussians per voxel, this induces a natural hierarchy: Gaussians are denser near the robot to preserve fine details, and sparser at large radii to reduce redundancy while maintaining rendering quality.

\textbf{Per-point attributes.}
For each valid 3D point, we associate appearance and geometric cues by sampling its RGB value and dense image feature at the corresponding pixel location, and by computing a normalized viewing direction from the camera center to the point using the known extrinsics. 
We then concatenate the 3D position, RGB, viewing direction, and the compressed feature vector to form a per-point attribute $\mathbf{a}_n$ used for subsequent spherical voxel aggregation.

\textbf{Anchor feature fusion within each voxel.}
For each spherical voxel $v$ (indexed by $(i_r,i_\theta,i_\phi)$), we collect the set of points
$\mathcal{P}_v=\{(\mathbf{x}_n,\mathbf{a}_n)\}$, where $\mathbf{a}_n$ includes the per-point appearance (RGB) and geometric attributes.
We define the anchor center as the mean position
\begin{equation}
\bar{\mathbf{x}}_v=\frac{1}{|\mathcal{P}_v|}\sum_{n\in\mathcal{P}_v}\mathbf{x}_n.
\label{eq:anchor_center}
\end{equation}

We aggregate per-point attributes using inverse-distance weights so that points closer to the anchor contribute more:
\begin{equation}
w_n=\frac{\left(\|\mathbf{x}_n-\bar{\mathbf{x}}_v\|_2+\epsilon\right)^{-1}}
{\sum_{m\in\mathcal{P}_v}\left(\|\mathbf{x}_m-\bar{\mathbf{x}}_v\|_2+\epsilon\right)^{-1}},
\qquad \sum_{n\in\mathcal{P}_v} w_n = 1.
\label{eq:invdist_weight}
\end{equation}
Following~\cite{pointnerf}, we optionally pass each neighbor's attribute through an MLP conditioned on its relative position, and then perform weighted pooling:
\begin{equation}
\tilde{\mathbf{a}}_n=\mathrm{MLP}\!\left(\big[\ \mathbf{a}_n\ \Vert\ (\mathbf{x}_n-\bar{\mathbf{x}}_v)\ \big]\right),
\qquad
\bar{\mathbf{a}}_v=\sum_{n\in\mathcal{P}_v} w_n\,\tilde{\mathbf{a}}_n.
\label{eq:anchor_feat_pool}
\end{equation}
Here $\big[\cdot \Vert \cdot\big]$ denotes vector concatenation, i.e., stacking the per-point attribute $\mathbf{a}_n$ and the relative offset $(\mathbf{x}_n-\bar{\mathbf{x}}_v)$ into a single feature vector before the MLP.
The fused anchor feature $\bar{\mathbf{a}}_v$ compactly summarizes the local appearance and geometry of voxel $v$ and is used as input to the subsequent sparse convolution and Gaussian decoding modules. Similarly, we compute the fused color $\bar{\mathbf{c}}_v\in\mathbb{R}^3$ by pooling the RGB entries in $\mathbf{a}_n$ (Eq.~\ref{eq:anchor_feat_pool}).

\textbf{Sparse convolution on the spherical grid.}
We pack voxel anchors into a sparse tensor indexed by $(i_r,i_\theta,i_\phi)$ and apply a sparse 3D CNN to propagate context across neighboring spherical voxels, producing refined voxel features.
These \emph{hierarchical spherical anchor features} are then used as inputs to our Gaussian decoding module, enabling compact Gaussian prediction for real-time rendering and streaming.

\subsection{Gaussian Parameters Prediction}
\label{sec:gaussian_pred}
Given the spherical voxel anchors, we decode a \emph{fixed} number of Gaussians per voxel, yielding a compact representation.
For each voxel $v$, we have the anchor center $\bar{\mathbf{x}}_v$, fused color $\bar{\mathbf{c}}_v$, and the spherical anchor feature after sparse convolution, denoted by $\mathbf{z}_v$.
We predict Gaussian parameters with a set of lightweight MLP heads
$\{h_{\Delta\mu},h_{\alpha},h_{s},h_{q},h_{\mathrm{dc}},h_{\mathrm{ho}}\}$.
This modular decoder decouples the prediction of Gaussian center (offset), visibility (opacity), geometry (scale/rotation), and appearance (spherical harmonic coefficients).
Each head takes the refined voxel feature $\mathbf{z}_v$ (optionally augmented with a scalar cue such as radius) and regresses the corresponding parameter group, keeping the decoder compact and efficient for real-time inference.

\textbf{Gaussian centers $\boldsymbol{\mu}$.}
We regress a bounded offset from the anchor center:
\begin{equation}
\boldsymbol{\mu}_v
=\bar{\mathbf{x}}_v+\gamma\big(2\sigma(\Delta\boldsymbol{\mu}_v)-1\big),
\qquad
\Delta\boldsymbol{\mu}_v = h_{\Delta\mu}(\mathbf{z}_v),
\label{eq:gs_mean}
\end{equation}
where $\gamma$ controls the maximum offset magnitude and the sigmoid bounds the offset to stabilize prediction.

\textbf{Opacity $\alpha$.} We predict opacity as $\alpha_v=\sigma\!\big(h_{\alpha}(\mathbf{z}_v)\big)$, where $\sigma(\cdot)$ maps it to $[0,1]$.

\textbf{Covariance $\Sigma$.}
We parameterize $\Sigma_v$ by per-axis scales and a unit quaternion:
\begin{equation}
\begin{aligned}
\mathbf{s}_v &= \kappa(r_v)\,\exp\!\Big(
\ell_{\min} + (\ell_{\max}-\ell_{\min})\,
\sigma\!\big(h_{s}(\mathbf{z}_v)\big)
\Big), \\
\mathbf{q}_v &= \mathrm{norm}\!\big(h_{q}(\mathbf{z}_v)\big),
\end{aligned}
\label{eq:gs_scale_quat}
\end{equation}

where $r_v=\|\bar{\mathbf{x}}_v\|_2$ is the anchor radius and $\kappa(r_v)$ is a distance-dependent scale factor that increases with $r_v$ so that farther voxels yield larger Gaussians; the scale head predicts bounded log-scales via a sigmoid mapping, and the rotation head outputs a quaternion normalized to enforce a valid rotation.
We then construct $\Sigma_v$ from $(\mathbf{s}_v,\mathbf{q}_v)$ following the standard 3DGS formulation.

\textbf{Appearance SH.}
We predict spherical-harmonic coefficients with two appearance heads.
The DC term is initialized from the fused RGB and refined by a residual predicted from $\mathbf{z}_v$, while higher-order terms are directly regressed from $\mathbf{z}_v$:
\begin{equation}
\mathbf{c}_{0,v} = \mathrm{RGB2SH}(\bar{\mathbf{c}}_v) + h_{\mathrm{dc}}(\mathbf{z}_v),
\qquad
\mathbf{c}_{>0,v} = h_{\mathrm{ho}}(\mathbf{z}_v).
\label{eq:gs_sh}
\end{equation}
This initialization anchors the predicted appearance to the observed color, while allowing the MLPs to learn view-dependent effects encoded in SH.

Since each voxel produces the same number of Gaussians, the radius-adaptive spherical voxelization (Sec.~\ref{sec:aggregation}) naturally yields denser Gaussians near the robot to preserve fine details, and sparser Gaussians in the far field to reduce redundancy while maintaining rendering quality.

\subsection{RobotPan Model Training}
\label{sec:model_train}

We train \textsc{RobotPan} with four objectives: two rendering losses for appearance supervision and two geometry losses for metric consistency.
At each time step $t$ and view $i$, we obtain a sparse set of LiDAR-supervised pixels by projecting the synchronized LiDAR scan into the $i$-th camera using calibration.
We denote this valid set as $\Omega_i^t$, and for each $p\in\Omega_i^t$ we have a metric 3D target point $\mathbf{x}^{t,\mathrm{lid}}_{i,p}$ in the camera coordinate, together with its depth $z^{t,\mathrm{lid}}_{i,p}$.
Meanwhile, we use $\pi^3$~\cite{pi3} to estimate a pseudo normal target $\mathbf{n}^{t}_{i,p}$ for normal supervision.

\textbf{Appearance supervision.}
We render each camera view from the predicted Gaussians and supervise the rendered images with a combination of MSE and LPIPS:
\begin{equation}
\mathcal{L}_{\mathrm{rgb}}^{\mathrm{mse}}=\big\|\hat{I}_i^t-I_i^t\big\|_2^2,
\qquad
\mathcal{L}_{\mathrm{rgb}}^{\mathrm{lpips}}=\mathrm{LPIPS}(\hat{I}_i^t, I_i^t),
\label{eq:train_rgb}
\end{equation}
where $\hat{I}_i^t$ is the differentiable rendering result for view $i$ at time $t$.

\textbf{Scale-aware point loss (LiDAR-sparse).}
Following the scale-invariant formulation, we align predicted point maps to metric LiDAR points, but only on the sparse set $\Omega_i^t$.
Unlike solving for an optimal $s^\star$, we treat $s$ as a learnable scalar (shared across views in a sequence/batch) and optimize it jointly with network parameters:
\begin{equation}
\mathcal{L}_{\mathrm{points}}
=
\frac{1}{\sum_{t,i}|\Omega_i^t|}
\sum_{t}\sum_{i=1}^{V}\sum_{p\in\Omega_i^t}
\frac{1}{z^{t,\mathrm{lid}}_{i,p}+\epsilon}\;
\big\| s\,\hat{\mathbf{x}}^{t}_{i,p}-\mathbf{x}^{t,\mathrm{lid}}_{i,p}\big\|_1,
\label{eq:train_points}
\end{equation}
where $\hat{\mathbf{x}}^{t}_{i,p}$ is the predicted 3D point from the point map $\mathbf{X}_i^t$ at pixel $p$.

\textbf{Normal loss.}
To encourage locally smooth surfaces, we supervise normals with the angular loss.
For each predicted point map $\mathbf{X}_i^t$, we compute the predicted normal $\hat{\mathbf{n}}^{t}_{i,p}$ at pixel $p$ from cross products of adjacent vectors on the image grid, and compare it with the pseudo ground-truth normal $\mathbf{n}^{t}_{i,p}$ estimated by $\pi^3$~\cite{pi3}:
\begin{equation}
\mathcal{L}_{\mathrm{normal}}
=
\frac{1}{\sum_{t,i}|\mathcal{V}_i^t|}
\sum_{t}\sum_{i=1}^{V}\sum_{p\in\mathcal{V}_i^t}
\arccos\!\big(\hat{\mathbf{n}}^{t}_{i,p}\cdot \mathbf{n}^{t}_{i,p}\big),
\label{eq:train_normal}
\end{equation}
where $\mathcal{V}_i^t$ denotes all valid pixels where normals can be computed (e.g., excluding boundaries/invalid depths).

\textbf{Overall objective.}
Our final training loss is a weighted sum of the four terms:
\begin{equation}
\mathcal{L}
=
\lambda_{\mathrm{mse}}\mathcal{L}_{\mathrm{rgb}}^{\mathrm{mse}}
+
\lambda_{\mathrm{lpips}}\mathcal{L}_{\mathrm{rgb}}^{\mathrm{lpips}}
+
\lambda_{\mathrm{pts}}\mathcal{L}_{\mathrm{points}}
+
\lambda_{\mathrm{nrm}}\mathcal{L}_{\mathrm{normal}}.
\label{eq:train_total}
\end{equation}

\begin{figure}[t]
  \centering
  \includegraphics[width=0.95\linewidth]{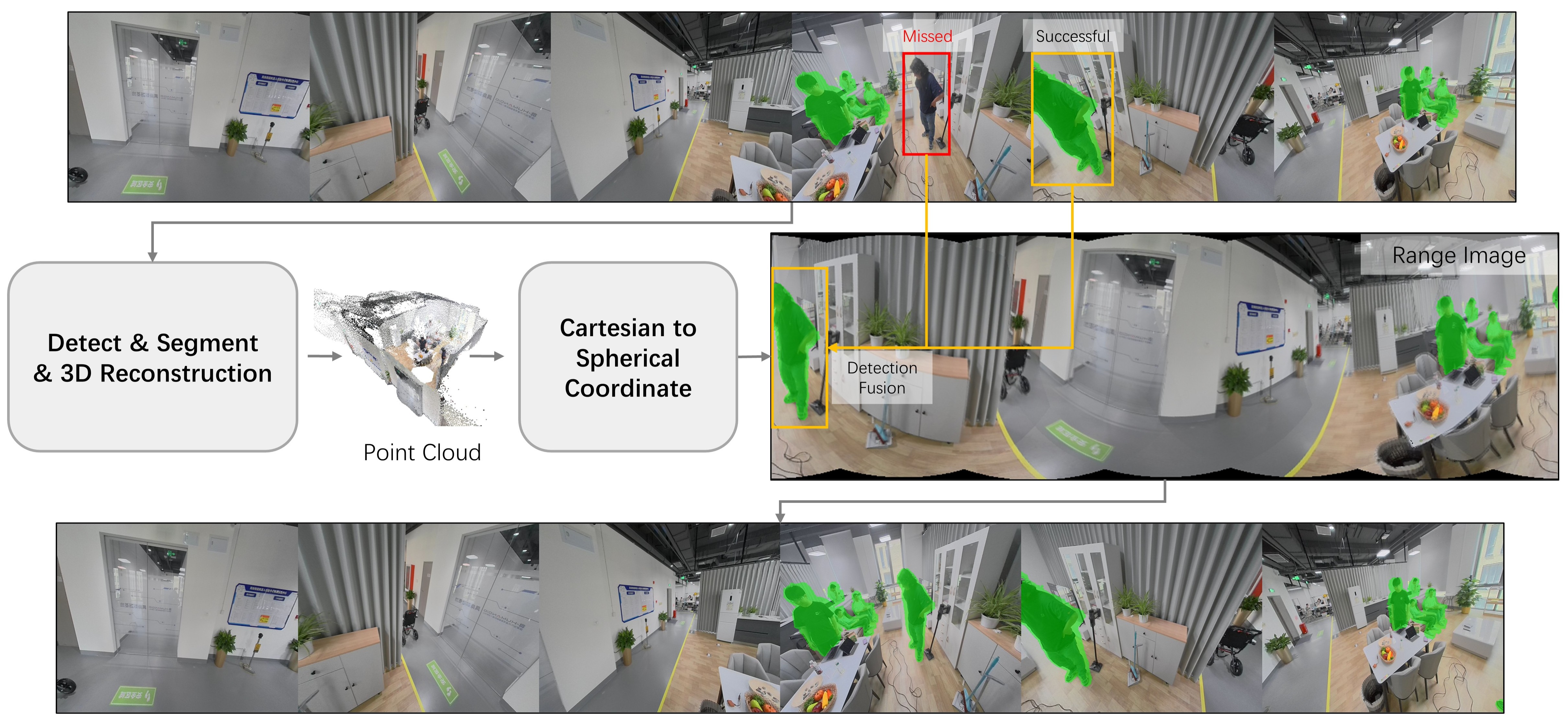}
  \caption{\textbf{Multi-view consistent dynamic region identification via range-image fusion.} We segment dynamic regions per view, reconstruct a shared 3D point cloud, project points to spherical coordinates to form a panoramic range image, and fuse multi-view results in the range-image domain to mitigate missed detections, yielding a robust dynamic-region mask for dynamic/static splitting. }
  \label{fig:dynamic_region_pipeline}
\end{figure}

\subsection{Streaming Fusion from Multi-Camera Sequences}
\label{sec:streaming}
\textsc{RobotPan} can predict a set of 3D Gaussians at every time step. A na\"{i}ve  way to reconstruct a long sequence is to concatenate the per-frame predictions. However, this quickly leads to (i) substantial redundancy, since many Gaussians repeatedly represent the same time-invariant scene content across frames, and (ii) visible artifacts at frame boundaries, where independently estimated Gaussian sets are not perfectly consistent.

\noindent\textbf{Streaming fusion.} We propose a streaming fusion strategy that maintains a compact, shared set of Gaussians for persistent scene content, and adds new Gaussians only when previously unseen or truly dynamic regions emerge.
Concretely, we represent the scene at time $t$ as
\begin{equation}
\mathcal{G}^{t} \;=\; \mathcal{G}_{\text{shared}} \;\cup\; \mathcal{G}^{t}_{\text{dyn}},
\end{equation}
where $\mathcal{G}_{\text{shared}}$ is a \emph{frame-shared} compact Gaussian set, and $\mathcal{G}^{t}_{\text{dyn}}$ captures time-dependent content specific to frame $t$. At each new time step, instead of re-adding all Gaussians, we \emph{only update} (i) the dynamic component and (ii) previously unseen regions. We first predict per-view motion/instance mask and then enforce multi-view consistency to reduce missed detections. Specifically, as shown in Fig.~\ref{fig:dynamic_region_pipeline}, for each camera we lift the per-pixel dynamic mask to 3D using the predicted point map, transform the resulting 3D points into spherical coordinates, and rasterize them into a range image; we then fuse all views by a pixel-wise \emph{union} operation in the spherical/range-image domain, producing a multi-view consistent dynamic region.
To identify unseen regions, we project the accumulated Gaussians from past frames into the current views and detect \emph{holes} where no prior content explains the observations; we fill these holes using the current frame's predicted point map and instantiate new Gaussians accordingly. 

To improve temporal coherence, we allow Gaussians from past frames to be \emph{gently refined} using the current observations via a lightweight tiny-MLP $f_{\theta}^{t}$~\cite{3dgstream, instantngp}. Specifically, we attach a per-frame tiny-MLP that predicts residual corrections for shared Gaussians (e.g., center offset/shape/opacity/appearance) conditioned on their 3D locations and color,
\begin{equation}
\Delta \mathbf{p} \;=\; f_{\theta}^{t}(\mathbf{x}, \mathbf{c}), \qquad
\tilde{\mathbf{p}}^{t} \;=\; \mathbf{p} \oplus \Delta \mathbf{p},
\end{equation}
where $\mathbf{p}$ denotes the parameters of a shared Gaussian, and $\oplus$ is the corresponding parameter composition (e.g., additive updates for position and log-scale, and quaternion composition for rotation). We optimize $\mathcal{G}_{\text{shared}}$, $\mathcal{G}^{t}_{\text{dyn}}$, and $f_{\theta}^{t}$ using multi-view rendering losses (photometric consistency, optionally depth consistency), with static-region masking when enforcing cross-frame constraints. This streaming update avoids redundant Gaussians and mitigates stitching artifacts, while remaining efficient for long sequences.

\textbf{Storage and reconstruction.} For compact storage, we save (i) a single shared Gaussian set $\mathcal{G}_{\text{shared}}$ reused by all frames, and (ii) for each time step $t$, the dynamic Gaussians $\mathcal{G}^{t}_{\text{dyn}}$ together with the tiny-MLP parameters $\theta^{t}$. To reconstruct frame $t$, we combine the refined shared Gaussians (through $f_{\theta}^{t}$) with the frame-specific dynamic Gaussians $\mathcal{G}^{t}_{\text{dyn}}$. A detailed comparison of storage requirements is provided in Sec.~\ref{exp:streaming_nvs}.
% The comparison of storage details in Sec.~\ref{exp:streaming_nvs}.

\section{Experiments}
We first introduce the design of our robot head, detailing the surround-view camera--LiDAR configuration and mounting geometry. Next, we describe our data collection protocol using a custom wearable rig. Building on this hardware and collection pipeline, we present the RobotPan dataset and its associated benchmark, covering dataset statistics, distribution, and experimental settings (Sec.~\ref{exp:setup}). We then evaluate and compare our approach against prior methods on three core tasks:
(i) \textbf{3D reconstruction / point-map estimation} (Sec.~\ref{exp:reconstruction}),
(ii) \textbf{sparse-view novel view synthesis} (Sec.~\ref{exp:nvs}), and
(iii) \textbf{streaming novel view synthesis} (Sec.~\ref{exp:streaming_nvs}).
Across all tasks, our method achieves state-of-the-art or comparable performance relative to existing feed-forward 3D reconstruction and generalizable novel view synthesis approaches on the RobotPan benchmark.

\begin{figure}[t]
  \centering
  \includegraphics[width=0.9\linewidth]{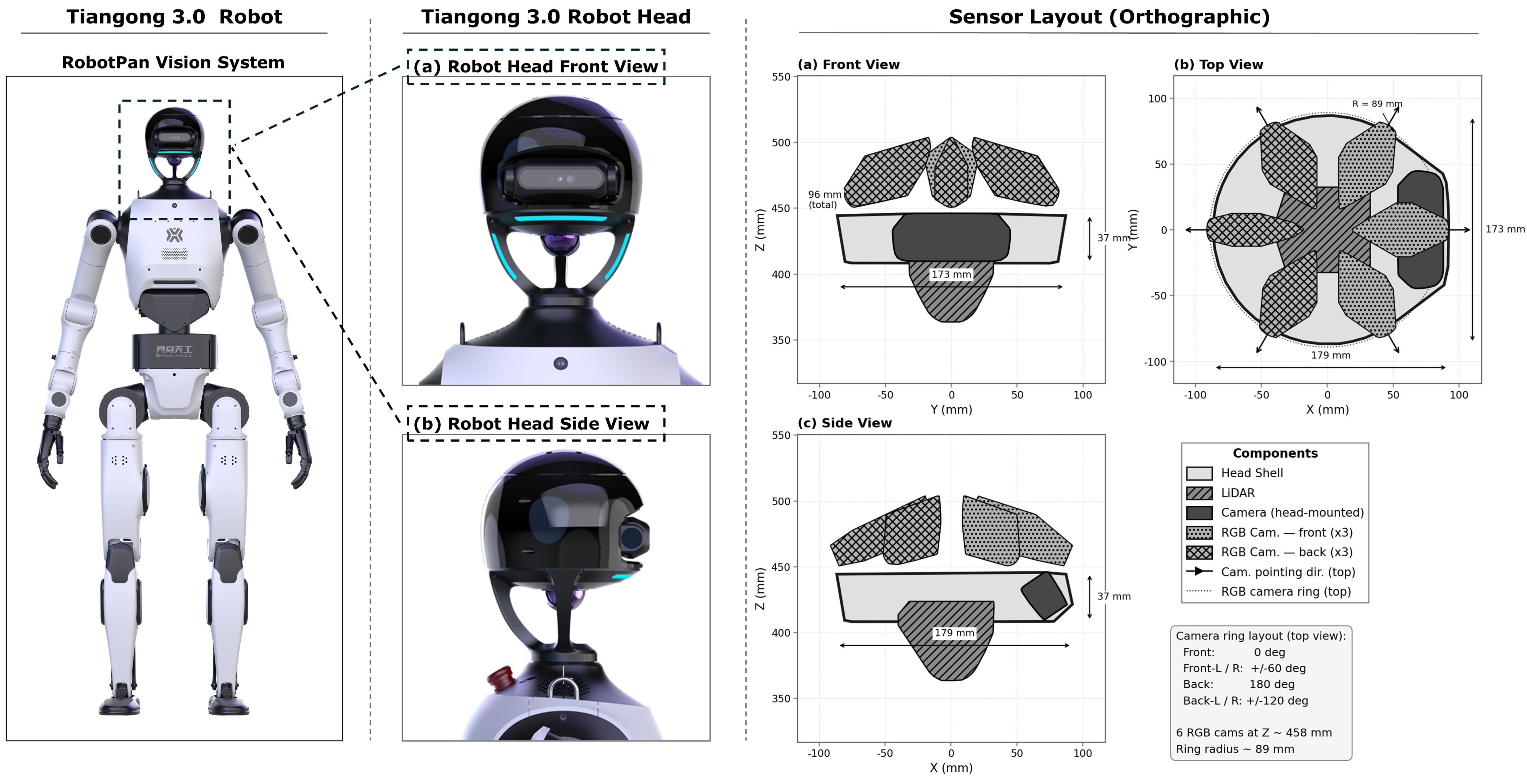}
\caption{\textbf{\emph{Tiangong 3.0}  robot head and its sensor layout.} \textbf{Left:} front and side views of the robot head. \textbf{Right:} orthographic views of the head-mounted sensing system, showing the arrangement of six RGB cameras and one LiDAR with annotated dimensions and viewing directions.}
  \label{fig:robothead}
\end{figure}

\begin{figure}[t]
  \centering
  \includegraphics[width=0.95\linewidth]{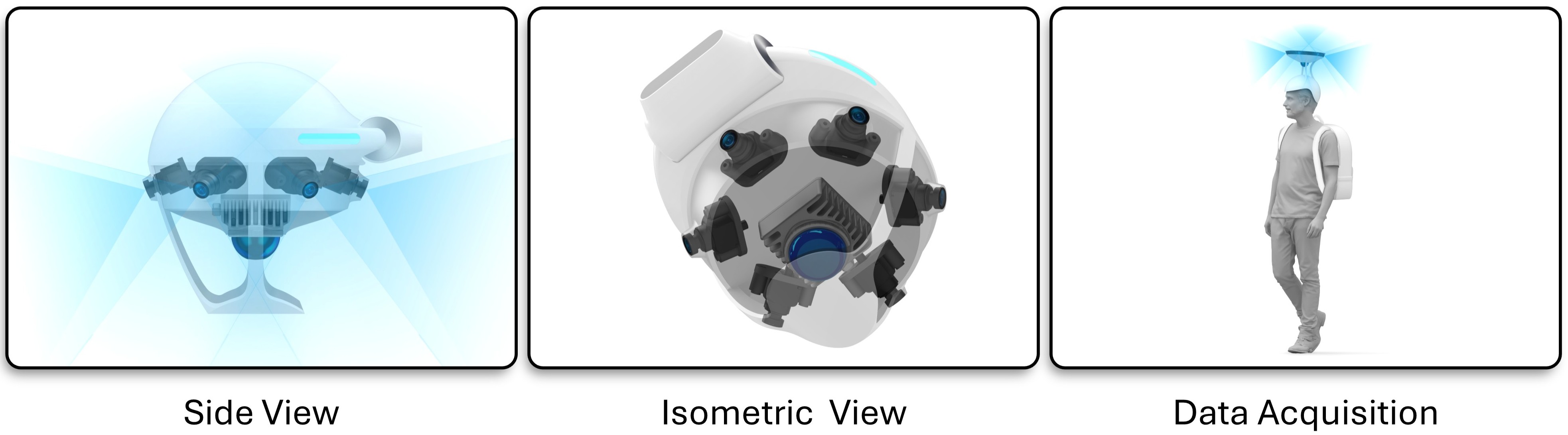}
\caption{\textbf{Sensor layout and data acquisition.} \textbf{Left and middle}: side and isometric views of our surround-view rig with six RGB cameras and a 40-beam LiDAR. \textbf{Right}: wearable data collection setup with sensor height matched to the humanoid robot.}
  \label{fig:sensor_layout}
\end{figure}

\subsection{Experimental Setup}
\label{exp:setup}

\textbf{Robotic sensor layout.}
\label{sensor_layout}
Our humanoid robotic platform, \emph{Tiangong 3.0}, is equipped with the custom-designed RobotPan vision system, featuring a highly integrated surround-view camera array coupled with a centrally mounted LiDAR sensor. As illustrated in Fig.~\ref{fig:robothead}, the visual perception system is engineered around a compact head shell, comprising six outward-facing RGB cameras arranged in a precise circular ring. To ensure optimal omnidirectional coverage with minimal occlusion, these cameras are uniformly distributed along a ring with an $89$\,mm radius at a consistent elevation. Specifically, they are positioned at $0^\circ$ (front), $\pm 60^\circ$ (front-left/right), $\pm 120^\circ$ (back-left/right), and $180^\circ$ (rear). Coupled with each camera's wide horizontal field of view of $118^\circ$ and vertical field of view of $92^\circ$, this geometry guarantees robust, multi-view visual overlap. Complementing the optical sensors, a 40-beam spinning LiDAR is mounted at the crown of the head shell. This LiDAR provides full $360^\circ$ horizontal coverage and a $59^\circ$ vertical field of view, functioning in tandem with the camera ring to deliver dense, multi-modal spatial data essential for downstream 3D modeling and accurate scene reconstruction.

\noindent\textbf{Data collection.}
To collect training and evaluation data under robot-realistic viewpoints, as shown in Fig.~\ref{fig:sensor_layout}, we design a head-mounted wearable capture system whose sensor layout matches the proposed robot surround-view configuration.
The wearable rig is optimized for human portability while preserving the camera--LiDAR geometry of the final robot installation.
Data are recorded by human operators (average height $\sim$160\,cm) wearing the rig on the head to match the sensor height of the target humanoid platform.
We further add a neck stabilizer to reduce head-induced motion and enforce conservative motion profiles during capture: walking speed is limited to $\leq 1.2$\,m/s and turning speed to $\leq 0.4$\,rad/s for data consistency.

\noindent\textbf{RobotPan dataset.}
Our dataset contains 339 synchronized clips, each with 200 frames, captured with a calibrated rig consisting of six RGB cameras and one LiDAR.
The sequences cover diverse \emph{indoor} environments (office buildings, households, exhibition halls, and factories) and \emph{outdoor} environments (urban blocks, industrial parks, and roads).
All sensors are time-synchronized and jointly calibrated, enabling consistent multi-view and multi-sensor geometry.
The RGB cameras record at a resolution of $1920{\times}1536$, and we downsample images to $518{\times}406$ during training.
The dataset features substantial non-rigid and dynamic content, with dynamic scenes accounting for approximately 80\% of the clips, including crowded pedestrian scenarios.
We split the data into 80\%/10\%/10\% for training/validation/test. In addition to the proposed data set, we introduce the task-specific dataset in the corresponding task section.

\noindent\textbf{Implementation details.}
We follow the VGGT-style~\cite{vggt} \emph{alternating-attention} design, interleaving view-wise (per-view) and global (across-view) self-attention; we use $36$ blocks in total ($18$ view-wise and $18$ global). We optimize with AdamW (lr $=5{\times}10^{-4}$, weight decay $=10^{-4}$), resize input images so that the longer side is at most $518$ pixels, and apply gradient-norm clipping with threshold 1.0 for training stability. Training is conducted in two stages: we initialize from the $\pi^3$ checkpoint, first fine-tune the geometry predictor with sparse LiDAR supervision and known camera calibration, and then enable the rendering component for joint training.  We complete the training on 8 GPUs. We leverage bfloat16 precision and gradient checkpointing to improve GPU memory and computational efficiency.

\begin{table}[!t]
\centering
\caption{Point map estimation results on the proposed dataset under sparse-view input, evaluated with and without camera poses.}
\label{tab:geo_eval_robotpan}
\renewcommand{\arraystretch}{1.12}
\resizebox{\columnwidth}{!}{%
\begin{tabular}{lcccccc}
\toprule
\multirow{2}{*}{\textbf{Method}}
& \multicolumn{2}{c}{Acc.$\downarrow$}
& \multicolumn{2}{c}{Comp.$\downarrow$}
& \multicolumn{2}{c}{Overall$\downarrow$} \\
\cmidrule(lr){2-3}\cmidrule(lr){4-5}\cmidrule(lr){6-7}
& w/o p. & w/ p. & w/o p. & w/ p. & w/o p. & w/ p. \\
\midrule
Dust3R~\cite{dust3r} & 0.783 & 0.634 & 0.525 & 0.418 & 0.654 & 0.521 \\
Fast3R~\cite{fast3r} & 0.716 & 0.562 & 0.482 & 0.350 & 0.602 & 0.460 \\
FLARE~\cite{flare}   & 0.660 & 0.518 & 0.421 & 0.298 & 0.549 & 0.404 \\
VGGT~\cite{vggt}     & \underline{0.611} & \underline{0.479} & 0.388 & \underline{0.245} & 0.496 & \underline{0.362} \\
Pi3~\cite{pi3}       & \textbf{0.580} & 0.440 & \underline{0.353} & 0.221 & \textbf{0.460} & 0.332 \\
\midrule
\textbf{Ours}        & 0.612 & \textbf{0.409} & \textbf{0.350} & \textbf{0.126} & \underline{0.482} & \textbf{0.268} \\
\bottomrule
\end{tabular}%
}
\end{table}

\begin{table}[!t]
\centering
\caption{Sparse-view reconstruction on DTU and ETH3D.}
\label{tab:geo_eval_dtu_eth3d}
\renewcommand{\arraystretch}{1.1}
\resizebox{\columnwidth}{!}{%
\begin{tabular}{lcccccc}
\toprule
\multirow{2}{*}{\textbf{Method}}
& \multicolumn{3}{c}{\textbf{DTU}}
& \multicolumn{3}{c}{\textbf{ETH3D}} \\
\cmidrule(lr){2-4}\cmidrule(lr){5-7}
& Acc.$\downarrow$ & Comp.$\downarrow$ & Overall$\downarrow$
& Acc.$\downarrow$ & Comp.$\downarrow$ & Overall$\downarrow$ \\
\midrule
Dust3R~\cite{dust3r}   & 3.8562 & 3.1219 & 3.4891 & 0.4856 & 0.7114 & 0.5984 \\
MASt3R~\cite{mast3r}   & 4.2380 & 3.2695 & 3.7537 & 0.3417 & 0.3626 & 0.3522 \\
Spann3R~\cite{spann3r} & 4.3097 & 4.5573 & 4.4335 & 1.1589 & 0.8005 & 0.9797 \\
VGGT~\cite{vggt}       & 3.5042 & 2.7254 & 3.1152 & 0.3127 & 0.2946 & 0.3456 \\
Pi3~\cite{pi3}         & \textbf{3.0213} & \textbf{2.1322} & \underline{3.0785} & \underline{0.1341} & \textbf{0.1169} & \textbf{0.1672} \\
\midrule
\textbf{Ours}          & \underline{3.3049} & \underline{2.2424} & \textbf{3.0251} & \textbf{0.1132} & \underline{0.1240} & \underline{0.1841} \\
\bottomrule
\end{tabular}%
}
\end{table}

\subsection{Point Map Estimation}
\label{exp:reconstruction}

\noindent\textbf{Datasets.}
We evaluate sparse-view geometry reconstruction on both our proposed surround-view robotic dataset and two widely used public benchmarks, DTU~\cite{dtu} and ETH3D~\cite{eth3d}. These datasets cover complementary reconstruction scenarios, including object-centric indoor captures and scene-level real-world environments. For DTU and ETH3D, we follow the protocol in~\cite{pi3} and sample keyframes every five images to construct sparse-view evaluation sequences.

\noindent\textbf{Metrics and baselines.}
We compare our method against recent feed-forward multi-view geometry reconstruction approaches, including Dust3R~\cite{dust3r}, Fast3R~\cite{fast3r}, FLARE~\cite{flare}, VGGT~\cite{vggt}, Pi3~\cite{pi3}, MASt3R~\cite{mast3r}, and Spann3R~\cite{spann3r}, depending on benchmark availability. Consistent with prior works~\cite{cut3r,pi3,vggt}, we report Accuracy (Acc.), Completeness (Comp.), and overall Chamfer distance (Overall), where lower is better for all metrics. Predicted point maps are aligned to the ground truth using the Umeyama algorithm for an initial Sim(3) alignment, followed by refinement with Iterative Closest Point (ICP).

\begin{figure}[t]
  \centering
  \includegraphics[width=0.95\linewidth]{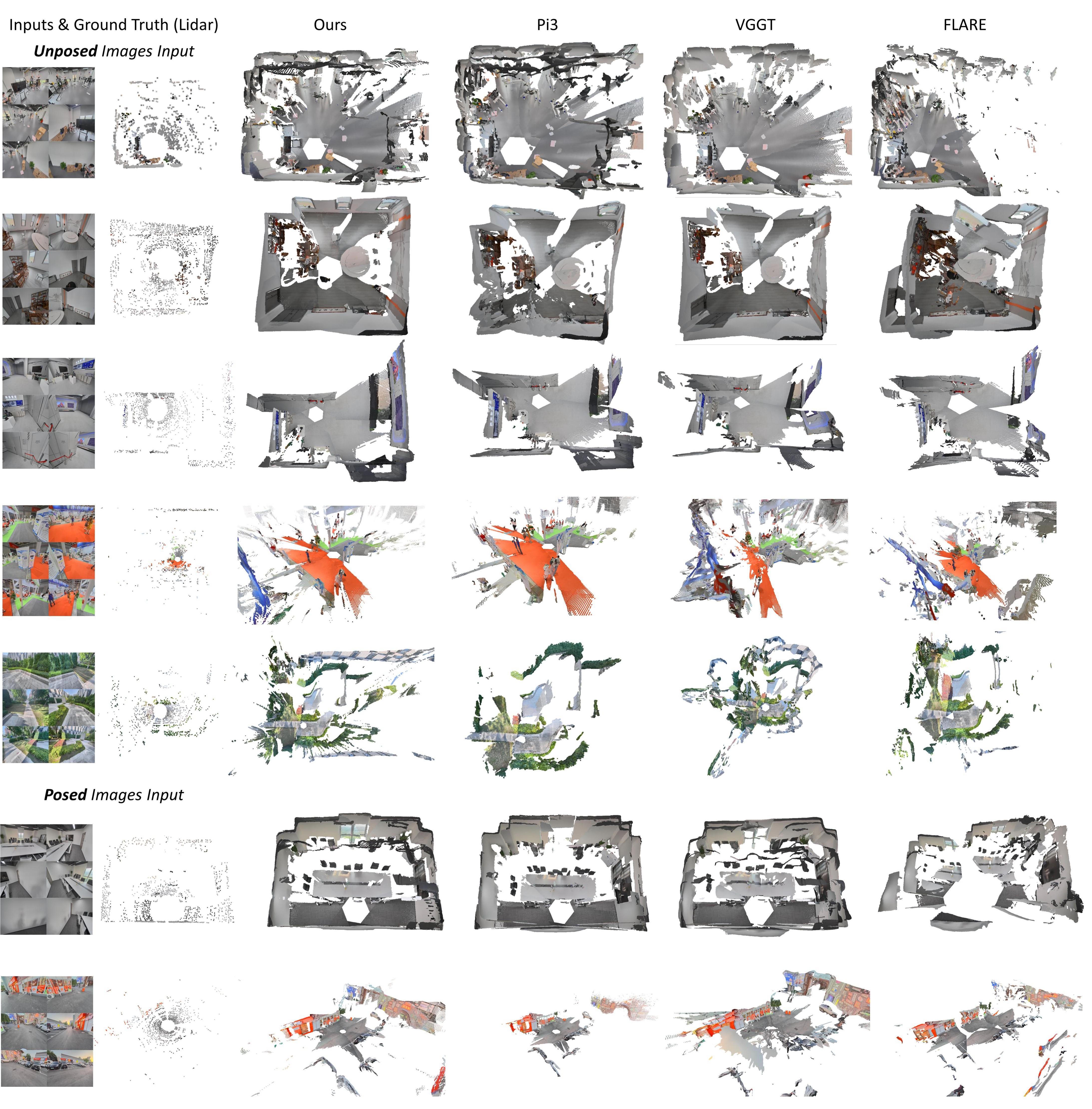}
\caption{\textbf{Qualitative comparison of feed-forward 3D reconstruction.} For each example, we show the input images, the LiDAR point cloud as ground truth, and the reconstructed geometry from different methods. Our method recovers more complete structures with sharper details.}
  \label{fig:recon_comp}
\end{figure}

\noindent\textbf{Comparison.}
Following the evaluation protocol of~\cite{pi3}, we assess the quality of reconstructed multi-view point maps on our proposed dataset under both posed and unposed settings. As described in Sec.~\ref{sensor_layout}, the six cameras are mounted on a circular ring and have only limited overlap in their fields of view, making reconstruction significantly more challenging than in standard forward-facing or object-centric capture setups. To reflect this difficulty, we consider two settings on our dataset: with camera poses and without camera poses. The quantitative results are reported in Tab.~\ref{tab:geo_eval_robotpan} and Tab.~\ref{tab:geo_eval_dtu_eth3d}. On our proposed robotic benchmark, the results demonstrate strong robustness under extremely sparse-view surround-view perception. On DTU and ETH3D, our method also generalizes well across both object-level and scene-level reconstruction scenarios, achieving competitive performance against existing approaches.

\begin{figure}[t]
  \centering
  \includegraphics[width=0.8\linewidth]{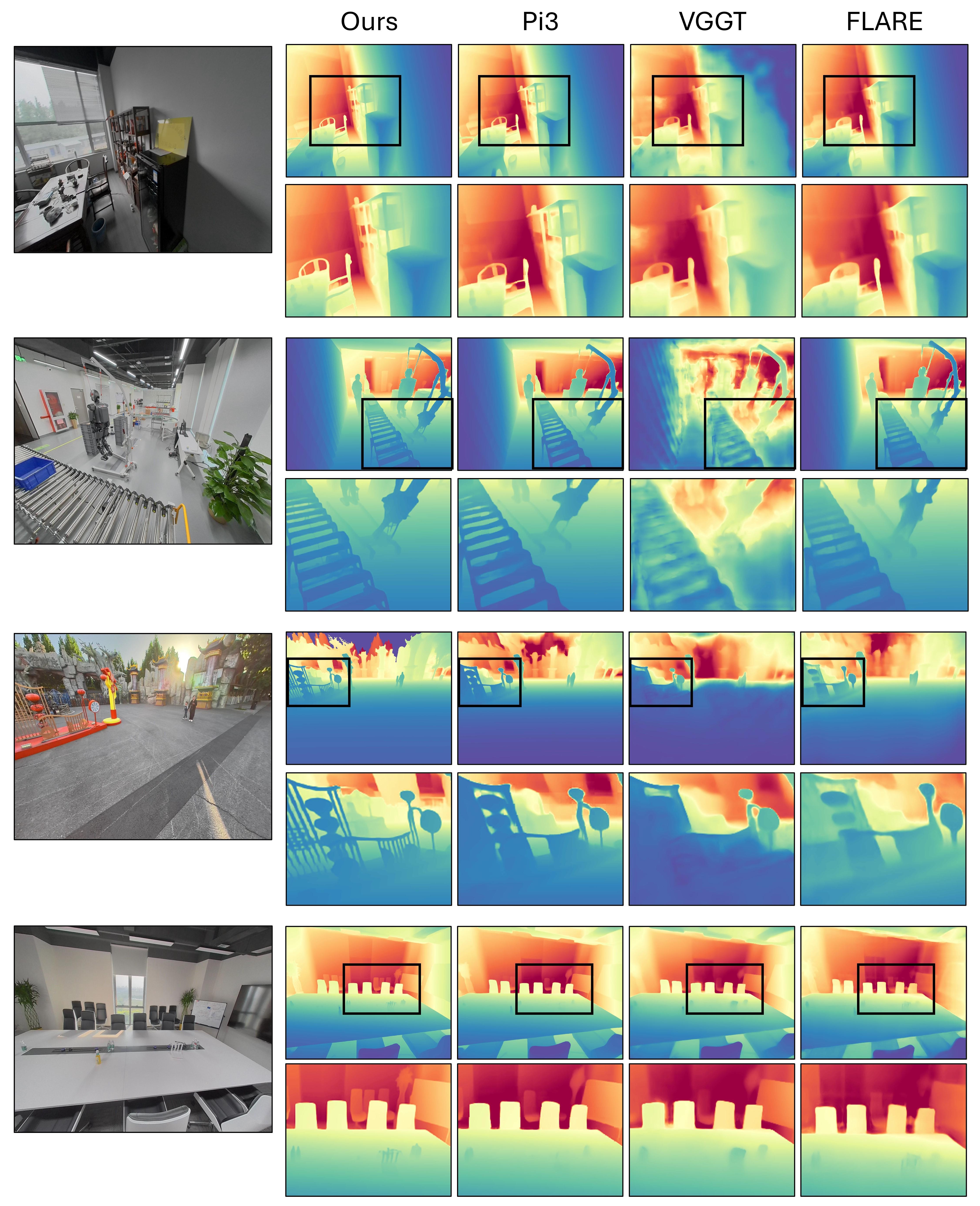}
\caption{\textbf{Qualitative comparison of monocular depth prediction.} For each example, we show the input image, predicted depth maps from different methods, and zoomed-in regions. Our method produces clearer boundaries and more consistent geometries across diverse scenes.}
  \label{fig:depth_comp}
\end{figure}

To provide a more comprehensive evaluation, we further analyze performance under sparse-view conditions. This setting, characterized by limited inter-frame overlap, is highly ill-posed and requires the model to exploit strong geometric and spatial priors. Despite this challenge, our method remains highly competitive in both global dense reconstruction (Fig.~\ref{fig:recon_comp}) and fine-grained geometric recovery, such as depth estimation (Fig.~\ref{fig:depth_comp}). These results indicate that the reconstruction module of RobotPan provides a strong geometric foundation for downstream robotic tasks, including mapping, localization, navigation, and fine-grained manipulation.

\begin{table}[t!]
\centering
\caption{Comparison with state-of-the-art generalized novel view synthesis methods on the proposed dataset. Our method achieves the best overall performance with the fewest number of Gaussians.}
\label{tab:nvs_eval_robotpan}
\renewcommand{\arraystretch}{1.05}
\small
\resizebox{\columnwidth}{!}{%
\begin{tabular}{lccccc}
\toprule
\textbf{Method} & \textbf{Paradigm} & \textbf{\#Gaussians}$\downarrow$ & \textbf{PSNR}$\uparrow$ & \textbf{SSIM}$\uparrow$ & \textbf{LPIPS}$\downarrow$ \\
\midrule
pixelSplat~\cite{pixelsplat} & Pixel-wise  & 3,783K & 17.79 & 0.447 & 0.443 \\
MVSplat~\cite{mvsplat}       & Pixel-wise  & 1,261K & 18.57 & 0.614 & 0.424 \\
FLARE~\cite{flare}           & Pixel-wise  & 1,261K & 18.29 & 0.625 & 0.393 \\
DepthSplat~\cite{depthsplat} & Pixel-wise  & 1,261K & \underline{22.97} & \underline{0.787} & \underline{0.200} \\
\midrule
\textbf{Ours}                & Spherical voxel-wise & \textbf{327K} & \textbf{24.70} & \textbf{0.811} & \textbf{0.197} \\
\bottomrule
\end{tabular}%
}
\end{table}

\subsection{Generalized Novel View Synthesis}
\label{exp:nvs}

\noindent\textbf{Datasets.}
We evaluate generalized novel view synthesis on both our proposed surround-view robotic dataset and two public benchmarks, DL3DV-Benchmarks~\cite{dl3dv} and RealEstate10K~\cite{re10k}. Our proposed dataset is designed for sparse-view robotic perception with large viewpoint changes and limited overlap across input cameras, making view synthesis substantially more challenging than standard forward-facing or object-centric settings. To further assess generalization, we also report results on DL3DV-Benchmarks and RealEstate10K, which are widely used benchmarks for feed-forward novel view synthesis.

\begin{table}[t]
\centering
\caption{Quantitative comparison with state-of-the-art novel view synthesis methods on the DL3DV-Benchmarks and RealEstate10K datasets.}
\label{tab:nvs_eval_dl3dv_re10k}
\renewcommand{\arraystretch}{1.15}
\resizebox{\columnwidth}{!}{%
\begin{tabular}{lcccccc}
\toprule
\multirow{2}{*}{\textbf{Method}}
& \multicolumn{3}{c}{\textbf{DL3DV-Benchmarks}}
& \multicolumn{3}{c}{\textbf{RealEstate10K}} \\
\cmidrule(lr){2-4}\cmidrule(lr){5-7}
& PSNR$\uparrow$ & SSIM$\uparrow$ & LPIPS$\downarrow$
& PSNR$\uparrow$ & SSIM$\uparrow$ & LPIPS$\downarrow$ \\
\midrule
pixelSplat~\cite{pixelsplat} & 16.55 & 0.456 & 0.480 & 25.89 & 0.858 & 0.142 \\
MVSplat~\cite{mvsplat}       & 18.13 & 0.559 & 0.393 & 26.39 & 0.869 & 0.128 \\
FLARE~\cite{flare}           & 18.89 & 0.591 & 0.352 & 27.39 & 0.873 & \textbf{0.107} \\
DepthSplat~\cite{depthsplat} & \underline{19.24} & \textbf{0.620} & \underline{0.322} & \textbf{27.47} & \underline{0.889} & 0.114 \\
\midrule
\textbf{Ours}                & \textbf{20.13} & \underline{0.616} & \textbf{0.318} & \underline{27.43} & \textbf{0.896} & \underline{0.109} \\
\bottomrule
\end{tabular}%
}
\end{table}

\noindent\textbf{Metrics and baselines.}
We compare our method against representative feed-forward novel view synthesis approaches, including pixelSplat~\cite{pixelsplat}, MVSplat~\cite{mvsplat}, FLARE~\cite{flare}, and DepthSplat~\cite{depthsplat}. Following standard practice, we report PSNR, SSIM, and LPIPS, where higher PSNR/SSIM and lower LPIPS indicate better rendering quality. In addition to image quality, we also compare the number of rendered Gaussians to evaluate representation compactness and rendering efficiency.

\noindent\textbf{Comparison.}
Tab.~\ref{tab:nvs_eval_robotpan} reports quantitative results on our proposed dataset under the 4-view setting. Despite the highly sparse-view configuration, our method achieves the best performance across all three image quality metrics while using a substantially more compact Gaussian representation. Our method achieves the best rendering quality while using a substantially more compact representation. Specifically, it uses only 327K Gaussians, which is \textbf{3.86$\times$ fewer} than recent one-Gaussian-per-pixel methods and \textbf{11.57$\times$ fewer} than pixelSplat (multiple-Gaussian-per-pixel). This result highlights the advantage of our anchor-wise representation in producing higher-quality renderings with significantly reduced redundancy.

To further evaluate cross-dataset generalization, we compare with prior methods on DL3DV-Benchmarks and RealEstate10K in Tab.~\ref{tab:nvs_eval_dl3dv_re10k}. Our method achieves the best PSNR and LPIPS on DL3DV-Benchmarks, and the best SSIM on RealEstate10K, while remaining competitive on the other metrics. These results demonstrate that the proposed representation and reconstruction-to-rendering pipeline generalize well beyond the robotic surround-view setting.

\begin{figure}
  \centering
  % \captionsetup{skip=2pt}
  \includegraphics[width=0.95\linewidth]{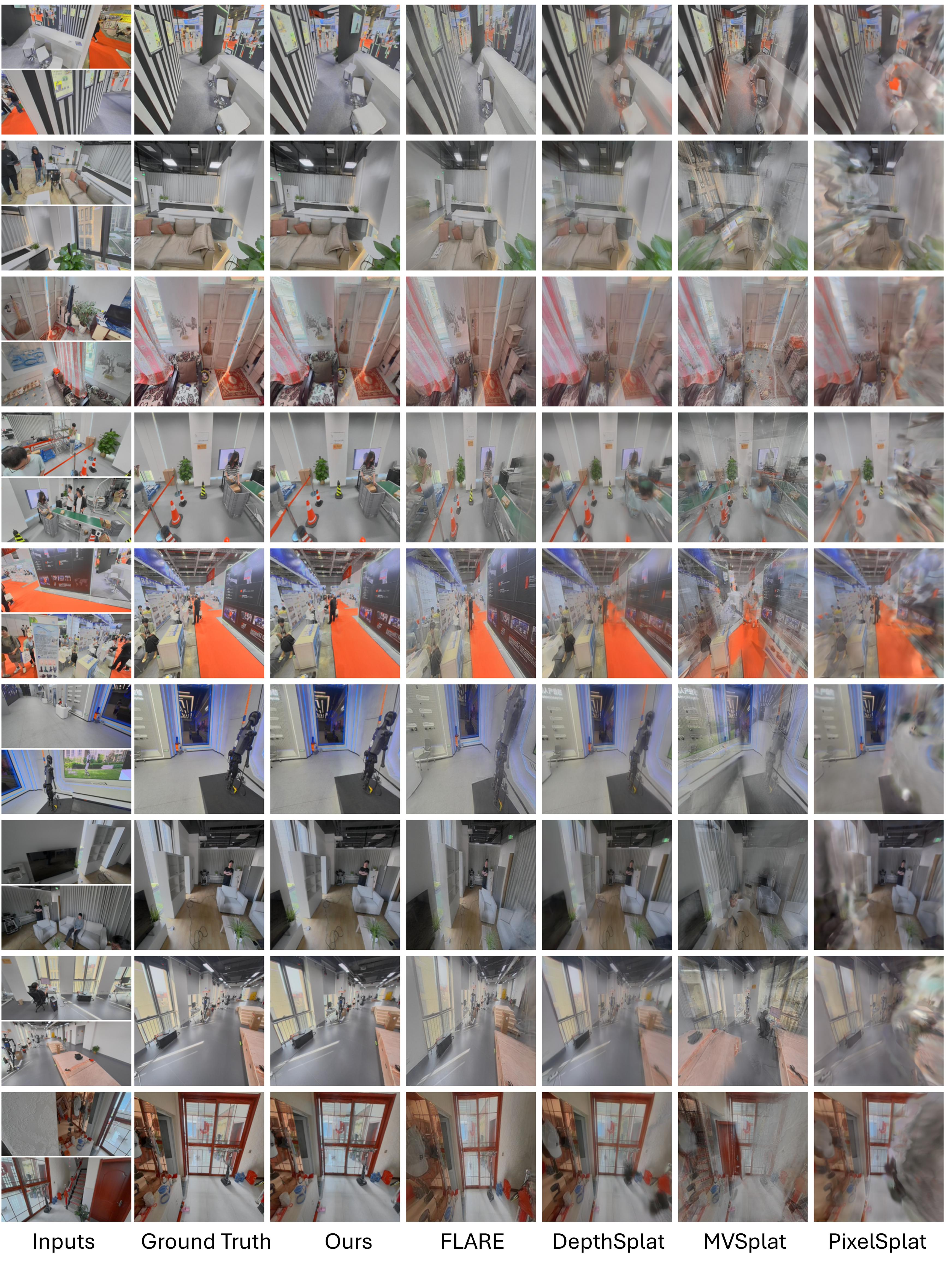}
  \caption{\textbf{Qualitative comparison of novel view synthesis results against state-of-the-art methods.} The first column shows the selected input reference views, and the remaining columns present the rendered novel views from different methods alongside the ground truth. Across challenging scenes, our method produces sharper structures, more faithful geometry, and fewer visual artifacts than competing approaches.}
  \label{fig:nvs_comp}
\end{figure}

\begin{figure}
  \centering
  % \captionsetup{skip=2pt}
  \includegraphics[width=0.95\linewidth]{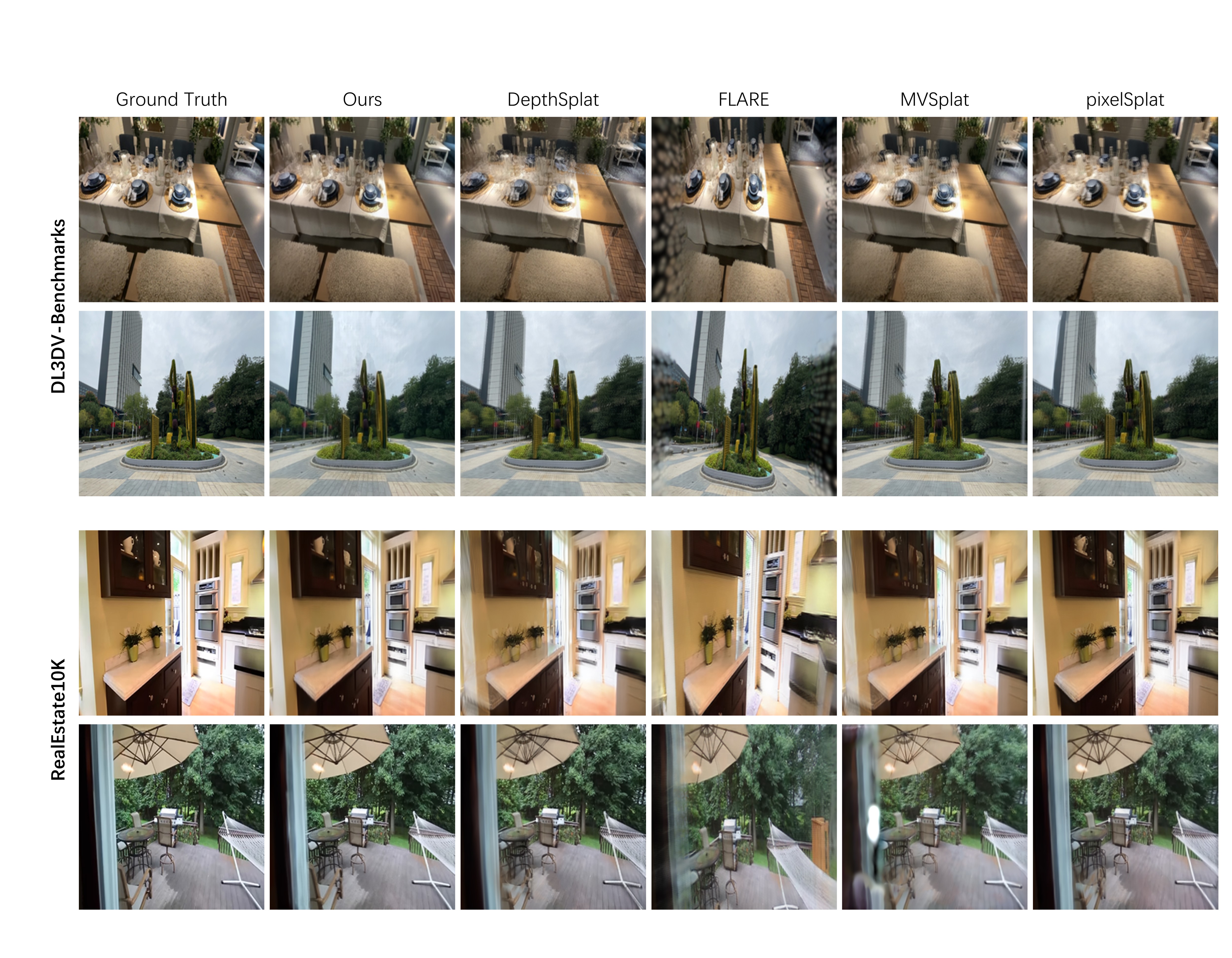}
     \caption{\textbf{Qualitative comparison on DL3DV-Benchmarks and RealEstate10K.} We show sparse input views and compare novel-view renderings from different methods against the ground truth. Our method preserves finer structures and cleaner boundaries.}
  \label{fig:nvs_dl3dv_re10k_comp}
\end{figure}

Qualitative comparisons are shown in Fig.~\ref{fig:nvs_comp} and Fig.~\ref{fig:nvs_dl3dv_re10k_comp}. Across diverse and challenging scenes, our method produces sharper structures, more faithful geometry, and fewer rendering artifacts than competing approaches. Together, these results show that RobotPan provides a strong solution for generalized novel view synthesis under sparse-view robotic perception settings and in general scenarios.

\subsection{Streaming Novel View Synthesis}
\label{exp:streaming_nvs}

\noindent\textbf{Datasets.}
We evaluate the performance of our streaming optimization framework on our collected multi-view dynamic scene dataset, and compare it with both offline and online training methods. Specifically, we select 10 sub-scenes from the captured data for comparison. Each sub-scene contains 200 frames, and each frame consists of 6 synchronized camera views, resulting in 1,200 images per scene. For evaluation, we randomly sample one view from each of the last 20 frames as the test set, while the remaining observations are used for training or streaming updates according to the protocol of each method.

\begin{table}[t]
\centering
\caption{Quantitative comparison of offline and online training methods on the proposed multi-view camera dynamic scene dataset.}
\label{tab:offline_online_comparison}
\renewcommand{\arraystretch}{1.15}
\resizebox{\columnwidth}{!}{%
\begin{tabular}{llcccc}
\toprule
\textbf{Category} & \textbf{Method} & \textbf{PSNR$\uparrow$} & \textbf{Train$\downarrow$} & \textbf{Render$\uparrow$} & \textbf{Storage$\downarrow$} \\
& & \textbf{(dB)} & \textbf{(s)} & \textbf{(FPS)} & \textbf{(MB)} \\
\midrule
\multirow{3}{*}{\textbf{Offline Training}} 
& Kplanes~\cite{kplanes}           & 23.17             & 52                    & 0.15                  & 1.5 \\
& 4DGS~\cite{4dgs}                 & 24.55             & 9.8                   & 30                    & \textbf{0.3} \\
& Spacetime-GS~\cite{spacetime_gs} & 25.41             & 51                    & 145                   & 0.8 \\
\midrule
\multirow{4}{*}{\textbf{Online Training}} 
& StreamRF~\cite{streamrf}         & 24.09             & 15.50                 & 8.3                   & 31.4 \\
& 3DGStream~\cite{3dgstream}       & 26.11             & 12.25                 & \underline{215}       & 7.8 \\
& IGS~\cite{IGS}                   & \underline{27.75} & 3.65                  & 204                   & \underline{6.5} \\
& Ours (Streaming)                   & \textbf{28.59}    & \textbf{0.47}         & \textbf{230}          & 7.2 \\
\bottomrule
\end{tabular}%
}
\end{table}

\noindent\textbf{Metrics and baselines.}
Following prior work~\cite{IGS}, we report \textbf{PSNR}, \textbf{Storage} usage, \textbf{Training} time, and \textbf{Rendering} speed for comparison with previous approaches. For \textit{online training} methods, the first-frame observations are used to initialize the scene representation, and the subsequent frames are incorporated incrementally through per-frame updates. In this setting, we compare against StreamRF~\cite{streamrf}, 3DGStream~\cite{3dgstream}, and IGS~\cite{IGS}. For \textit{offline training} methods, all frames in each sequence are jointly used for training. The offline baselines include K-Planes~\cite{kplanes}, 4DGS~\cite{4dgs}, and Spacetime-GS~\cite{spacetime_gs}. All reported metrics are averaged over the full 200-frame sequence, including frame 0.

\noindent\textbf{Comparison with offline and online methods.}
We compare our streaming framework against both offline and online training methods on the proposed multi-view camera dynamic scene dataset, with quantitative results reported in Tab.~\ref{tab:offline_online_comparison}. Compared to offline methods, our approach achieves the best rendering quality, reaching 28.59 dB PSNR and outperforming K-Planes, 4DGS, and Spacetime-GS by a clear margin. At the same time, our method preserves real-time rendering efficiency at 230 FPS, matching the fastest online baseline and substantially exceeding most offline alternatives. This efficiency is enabled by our feed-forward streaming design, which performs both scene initialization and subsequent updates without costly per-scene iterative optimization. In addition, thanks to our spherical voxelization scheme, the scene representation allocates Gaussians more efficiently across space, allowing us to model the scene with a smaller number of Gaussians while preserving reconstruction fidelity. This compact representation further reduces computational rendering overhead and contributes to the high rendering speed achieved by our proposed method. Although offline methods such as 4DGS are more compact in storage, they require sequence-level optimization over the entire scene and therefore are not suitable for practical low-latency streaming applications in robotics.

Compared to prior online training methods, our method achieves the best overall trade-off between rendering quality, training efficiency, rendering speed, and storage usage. Specifically, our streaming variant outperforms StreamRF, 3DGStream, and IGS in PSNR, while reducing the per-frame training time to only 0.47 seconds. This is more than $26\times$ faster than 3DGStream, more than $33\times$ faster than StreamRF, and about $7.8\times$ faster than IGS. In addition, our method achieves the highest rendering speed among all online baselines at 230 FPS. While IGS is slightly more compact in memory footprint, our method provides better rendering quality and substantially faster rendering, leading to the strongest overall performance in streaming settings.

These results demonstrate the effectiveness of our streaming design. By performing scene initialization and sequential updates in a feed-forward manner, our method avoids the heavy optimization cost of prior online approaches and thus greatly reduces update latency. Furthermore, our spherical voxelization scheme enables a compact Gaussian representation that allocates model capacity more efficiently in space, thereby lowering rendering cost while preserving reconstruction quality. Consequently, our method effectively overcomes the typical trade-off between quality and efficiency in online novel view synthesis.

\begin{table}[t!]
\centering
\caption{Ablation of Gaussian prediction paradigms on the proposed dataset. The default spherical voxel-wise prediction achieves the best trade-off between compactness and rendering quality.}
\label{tab:ablation_paradigm}
\renewcommand{\arraystretch}{1.05}
\small
\resizebox{\columnwidth}{!}{%
\begin{tabular}{lccccc}
\toprule
\textbf{Paradigm} & \textbf{\#Gaussians} & \textbf{Storage (MB)}$\downarrow$ & \textbf{PSNR}$\uparrow$ & \textbf{SSIM}$\uparrow$ & \textbf{LPIPS}$\downarrow$ \\
\midrule
Pixel-wise            & 1261K  & 378 & 21.43    & 0.713    & 0.341    \\
Voxel-wise            & \underline{439K}   & \underline{130} & \underline{23.12}    & \underline{0.791}    & \underline{0.236}    \\
Spherical voxel-wise  & \textbf{327K}  & \textbf{97} & \textbf{24.70} & \textbf{0.811} & \textbf{0.197} \\
\bottomrule
\end{tabular}%
}
\end{table}

\begin{table}[t]
\centering
\caption{Ablation study of key design choices on the proposed dataset for streaming novel view synthesis over 200 evaluation frames.}
\label{tab:ablation_modules}
\renewcommand{\arraystretch}{1.15}
\resizebox{\columnwidth}{!}{%
\begin{tabular}{lccccc}
\toprule
\textbf{Configuration} & \textbf{\#Gaussians} & \textbf{Storage (MB)}$\downarrow$ & \textbf{PSNR}$\uparrow$ & \textbf{SSIM}$\uparrow$ & \textbf{LPIPS}$\downarrow$ \\
 
\midrule
Naive Concatenation (w/o MLP)        & 65,400K & 19,400 & 24.66 & 0.806 & \underline{0.201} \\
w/o Range-Image Fusion      & \underline{3,094K} & 1,371 & \underline{24.71} & \underline{0.812} & 0.213 \\
w/o Tiny-MLP Refinement     & \textbf{3,023K} & \textbf{872} & 24.32 & 0.739 & 0.241 \\
\textbf{Full Model}         & \textbf{3,023K} & \underline{1,296} & \textbf{25.12} & \textbf{0.832} & \textbf{0.184} \\
\bottomrule
\end{tabular}%
}
\end{table}

\subsection{Ablation Study}
\noindent\textbf{Spherical voxel-wise prediction.}
Tab.~\ref{tab:ablation_paradigm} compares three Gaussian prediction paradigms: pixel-wise, Cartesian voxel-wise, and spherical voxel-wise. Both voxel-wise variants outperform pixel-wise prediction by aggregating nearby observations into compact anchor features, which reduces redundancy and improves rendering quality. In particular, spherical voxel-wise prediction achieves the best overall results, reducing the representation from 1261K to 327K Gaussians and from 378 MB to 97 MB, while improving PSNR/SSIM/LPIPS from 21.43/0.713/0.341 to 24.70/0.811/0.197. These results show that structured local aggregation is more suitable for our multi-view robotic setting than directly regressing dense pixel-aligned primitives.

Compared with Cartesian voxels, spherical voxels are better aligned with our robot-centric surround-view setting. Their volume increases with radius, naturally assigning finer resolution to near-field regions and coarser resolution to distant areas. This radius-adaptive hierarchy better matches the geometry distribution of embodied scenes, yielding both higher compactness and better rendering quality than uniform Cartesian voxelization.

\noindent\textbf{Streaming fusion.}
Tab.~\ref{tab:ablation_modules} validates the effectiveness of the proposed streaming fusion design for streaming novel view synthesis. Naive concatenation causes severe redundancy across time, increasing the representation to 65,400K Gaussians and 19,400 MB storage, which is impractical for long sequences despite reasonable rendering quality. This highlights the need to maintain a compact shared representation instead of directly accumulating frame-wise predictions.

Removing multi-view range-image fusion also reduces performance. Without this component, dynamic regions are identified independently in each view, making the system more vulnerable to missed detections and cross-view inconsistencies. As a result, the quality of streaming updates becomes less reliable, leading to lower PSNR/SSIM and higher LPIPS than the full model.

Removing the tiny-MLP refinement further degrades rendering quality, even with the same number of Gaussians as the full model. This shows that the gain does not come from using more primitives, but from better temporal refinement of the shared representation. Overall, the full model achieves the best trade-off between compactness and streaming rendering quality.
% \section{Conclusion}
% \label{sec:conclusion}
% We introduced a surround-view robotic vision system that integrates a six-camera rig with LiDAR to provide real-time 360$^\circ$ perception for embodied deployment, particularly under human-in-the-loop operation.
% To meet the geometric and systems constraints of teleoperation, data collection, and emergency takeover, we presented \textsc{RobotPan}, a feed-forward framework that predicts \emph{metric-scaled} and \emph{compact} 3D Gaussian primitives from calibrated sparse views, enabling real-time rendering, reconstruction, and streaming.
% We propose spherical representation with hierarchical spherical voxel priors, which naturally allocates higher capacity to near-field regions and reduces redundancy in the far field while preserving rendering fidelity.
% We further proposed a streaming fusion strategy that incrementally consolidates multi-frame, multi-camera predictions into a unified 3DGS, continuously updating dynamic content and preventing unbounded growth in static regions via selective appearance updates.
% To support future research, we released the first multi-sensor dataset tailored to 360$^\circ$ novel view synthesis and metric 3D reconstruction in robotics across navigation, manipulation, and locomotion scenarios.
% Extensive experiments showed that \textsc{RobotPan} achieves competitive quality against prior feed-forward reconstruction and view-synthesis methods while producing substantially fewer Gaussians, making it practical for real-time embodied deployment.

\section{Conclusion}
\label{sec:conclusion}
We introduced a surround-view robotic vision system that integrates a six-camera rig with LiDAR to provide real-time 360$^\circ$ perception for embodied deployment.
We presented \textsc{RobotPan}, a feed-forward framework that predicts \emph{metric-scaled} and \emph{compact} 3D Gaussian primitives from calibrated sparse views, enabling real-time rendering, reconstruction, and streaming.
Its hierarchical spherical voxel priors naturally allocate higher capacity to near-field regions and reduce redundancy in the far field, while a streaming fusion strategy incrementally consolidates multi-frame, multi-camera predictions into a unified 3DGS, updating dynamic content and preventing unbounded growth in static regions.
We also released the first multi-sensor dataset tailored to 360$^\circ$ novel view synthesis and metric 3D reconstruction in robotics, covering navigation, manipulation, and locomotion scenarios.
Extensive experiments show that \textsc{RobotPan} achieves competitive quality against prior methods while producing substantially fewer Gaussians, making it practical for real-time embodied deployment.

% \ifCLASSOPTIONcompsoc
%   \section*{Acknowledgments}
% \else
%   \section*{Acknowledgment}
% \fi
% \input{content/07_acknowledgements}

% \section*{Supplementary Information}
% \input{content/06_supplementary}

% \input{content/08_declaration}

% \appendices
% \input{content/09_appendix}

\bibliographystyle{IEEEtran}
\bibliography{sn-bibliography}

\end{document}